\documentclass[11pt]{article}

\usepackage[final]{acl}

\usepackage{times}
\usepackage{latexsym}

\usepackage[T1]{fontenc}
\usepackage[utf8]{inputenc}

\usepackage{microtype}

\usepackage{inconsolata}

\usepackage{graphicx}

\usepackage{amsmath,amssymb,amsfonts}
\usepackage[normalem]{ulem}
\useunder{\uline}{\ul}{}
\usepackage{booktabs}
\usepackage{multirow}
\usepackage{xspace}
\usepackage{xcolor}
\usepackage[most]{tcolorbox}

\newcommand{\Method}{N-GRPO\xspace}

\definecolor{promptboxframecolor}{HTML}{A9D18E}
\newtcolorbox{promptbox}[1][]{
  colback=promptboxframecolor!15!white,
  colframe=promptboxframecolor,
  fonttitle=\bfseries,
  boxrule=1pt,
  arc=2mm,
  toptitle=4pt, bottomtitle=1pt,
  left=8pt,right=8pt,top=12pt,bottom=12pt,
  enhanced,
  breakable,
  #1
}

\title{\Method: Embedding-Level Neighbor Mixing for Enhanced Policy Optimization}

\author{
 \textbf{Xukun Zhu\textsuperscript{1,2}},
 \textbf{Hang Yu\textsuperscript{2*}},
 \textbf{Peng Di\textsuperscript{2*}},
 \textbf{Linchao Zhu\textsuperscript{1*}}
\\
 \textsuperscript{1}Zhejiang University,
 \textsuperscript{2}Ant Group
\\
 \texttt{\{zhuxukun, zhulinchao\}@zju.edu.cn}
\\
 \texttt{\{hyu.hugo, dipeng.dp\}@antgroup.com}
}

\begin{document}
\maketitle

\begingroup
\renewcommand{\thefootnote}{*}
\footnotetext{Corresponding author.}
\endgroup

\begin{abstract}
% The effectiveness of Reinforcement Learning (RL) in reasoning tasks hinges on the policy's ability to explore diverse solution paths. However, conventional exploration strategies are constrained by discrete token-level sampling, which often restricts generated variations to surface-level lexical changes rather than meaningful semantic shifts. This inefficiency limits the model from discovering novel reasoning strategies, resulting in wasted exploration budget on redundant trajectories. To address this limitation, we propose \textbf{\Method}, a novel embedding-level exploration strategy integrated into the Group Relative Policy Optimization (GRPO) framework. Unlike traditional methods that scramble surface tokens, \Method performs exploration in the continuous embedding space via Semantic Neighbor Mixing. Specifically, during the rollout phase, we dynamically interpolate the embedding of a selected token with its nearest semantic neighbors in the vocabulary. This mechanism injects controllable, meaning-preserving perturbations that encourage the model to traverse diverse reasoning paths without sacrificing linguistic coherence. Extensive experiments on DeepSeek-R1-Distill-Qwen models across different scales demonstrate that \Method significantly outperforms standard baselines on challenging mathematical benchmarks and exhibits superior generalization on out-of-distribution scientific reasoning tasks.

The success of Large Language Models in mathematical reasoning relies heavily on the generation of diverse and valid solution paths during the rollout phase. However, current rollout techniques face a fundamental trade-off: token-level sampling often yields redundant trajectories that differ only in rephrasing, while embedding-level methods utilizing random noise frequently disrupt semantic consistency. To resolve this, we introduce \textbf{\Method}, a novel exploration strategy integrated into the Group Relative Policy Optimization (GRPO) framework. Rather than relying on token-level sampling or native embedding-level noise, our approach leverages Semantic Neighbor Mixing. This mechanism dynamically constructs input representations by mixing the embeddings of an anchor token and its nearest semantic neighbors, thereby injecting diversity while strictly adhering to the local semantic manifold. Experimental evaluations on the DeepSeek-R1-Distill-Qwen models across different sizes show that \Method not only achieves consistent improvements over strong baselines on math reasoning benchmarks but also exhibits robust generalization capabilities on out-of-distribution tasks.
\end{abstract}

\section{Introduction}
\label{sec:introduction}

Large Language Models (LLMs) have recently achieved remarkable progress on challenging reasoning tasks, especially in mathematical problem solving~\cite{guo2025deepseek,jaech2024openai,yang2025qwen3}. Reinforcement learning methods have emerged as an effective paradigm for enhancing LLM reasoning via improved trajectory exploration and policy optimization during training~\cite{ouyang2022training,zheng2025group}. Among them, GRPO~\cite{shao2024deepseekmath} is a representative approach that has demonstrated notable gains on multiple reasoning tasks.

High-quality exploration during the rollout phase is critical for the success of GRPO~\cite{wang2025hint}, yet token-level sampling methods~\cite{nguyen2024turning,bai2025online} often restrict exploration to simple paraphrasing and commutative re-orderings (e.g., ``1+2'' vs.\ ``2+1''). Since the underlying reasoning logic remains unchanged, these approaches result in the generation of redundant trajectories.

\begin{figure}[ht]
  \centering
  \includegraphics[width=\linewidth]{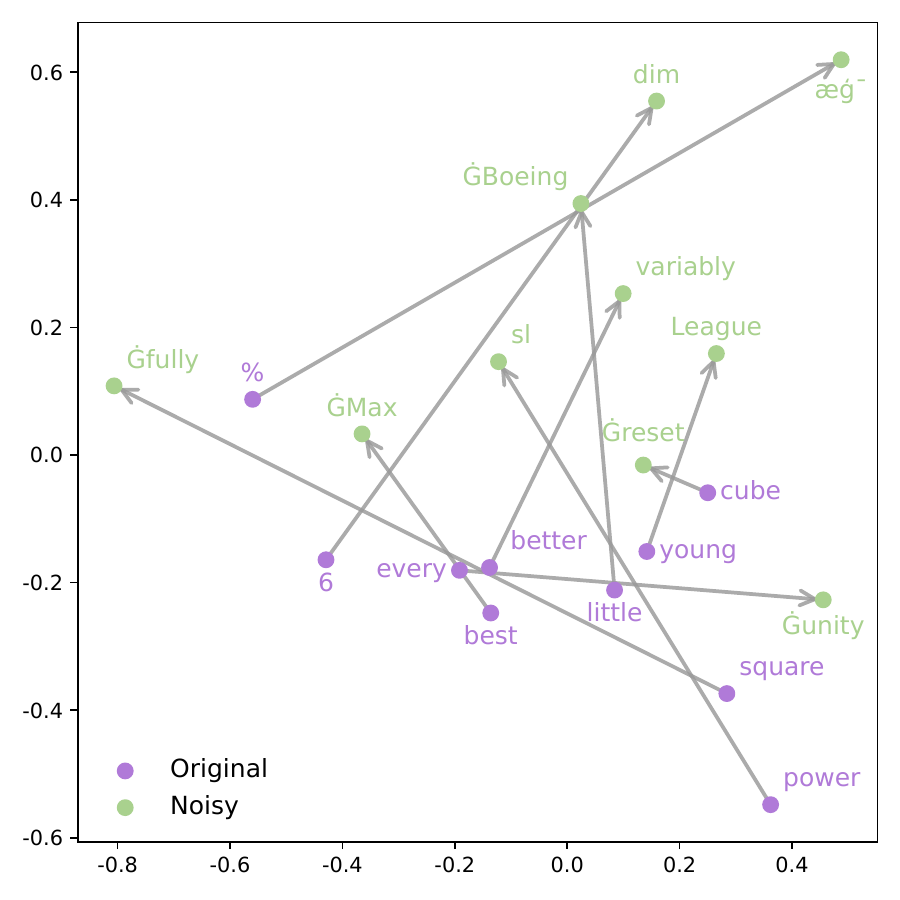}
  \caption{\textbf{Visualization of Semantic Drift caused by Unconstrained Noise.} We selected 10 random tokens (mathematical symbols and common words), added Gaussian noise to their embeddings, and projected them back to the nearest vocabulary token. The \textcolor[HTML]{B07AD8}{purple points} represent the original tokens, and the \textcolor[HTML]{A9D18E}{green points} represent the noise tokens. Lines connect the original token to its noisy projection. Observe that unconstrained noise causes significant semantic shifts, justifying the need for a semantically grounded exploration strategy.}
  \label{fig:noise_vis}
\end{figure}

To address this limitation, recent works inspired by Soft Thinking~\cite{zhang2025soft} attempt to enhance GRPO by conducting rollouts within the continuous embedding space. Since continuous representations lack inherent randomness, existing methods primarily follow two directions. HRPO~\cite{yue2025hybrid} extends Soft Thinking~\cite{zhang2025soft} into GRPO by feeding mixed continuous representations during rollout, but its randomness still ultimately stems from sampling discrete tokens, so its exploration is essentially discrete in nature and continues to yield redundant trajectories. Alternatively, ~\citet{butt2025soft} applies direct Gaussian noise injection to the embeddings or logits. However, such noise injection frequently disrupts semantic consistency. To demonstrate this, we conducted a preliminary experiment by injecting isotropic Gaussian noise into the embeddings of 10 randomly sampled tokens, consisting primarily of mathematical symbols and common function words, and decoding the resulting vectors to their nearest neighbors in the vocabulary. The resulting spatial displacement is visualized using Principal Component Analysis (PCA) in Figure~\ref{fig:noise_vis}. The projection demonstrates that this naive perturbation frequently pushes representations off the semantic manifold. Specifically, random noise frequently transforms the original tokens into unrelated tokens, leading to derailing the rollout trajectories.

\begin{figure*}[ht]
  \centering
  \includegraphics[width=\linewidth]{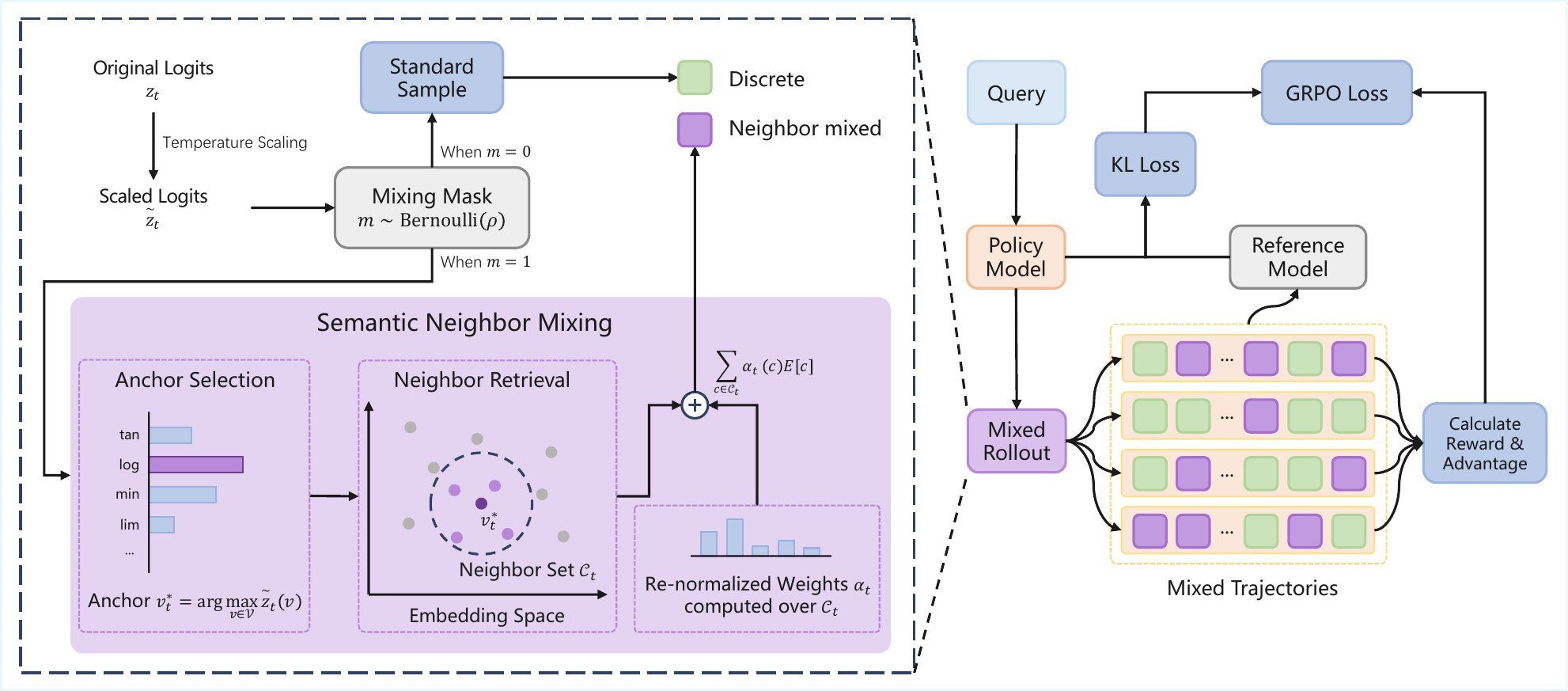}
  \caption{\textbf{Illustration of the \Method framework.} Left: Semantic Neighbor Mixing. The generation process is controlled by a mixing mask. When activated, the model bypasses discrete sampling and instead constructs a continuous embedding by aggregating a semantic neighbor set around an anchor token, using re-normalized weights. Right: Adaptation to GRPO. The proposed module is integrated into the rollout phase, where the policy model generates mixed trajectories containing both discrete tokens and neighbor-mixed embeddings.}
  \label{fig:pipeline}
\end{figure*}

We attribute this failure to the structure of the embedding space. Recent studies have demonstrated that the embedding space of Transformer models typically exhibits strong anisotropy~\cite{gao2018representation}. Embedding-level exploration during rollout must be adaptive to the local semantic context. To address this, we propose constructing an embedding-level perturbation by mixing the embedding of the sampled token with those of its nearest neighbor tokens. Since these neighbors are retrieved based on embedding similarity, they inherently cluster along the same semantic direction as the anchor token. Consequently, the mixed embedding remains confined within a valid semantic region, while providing the necessary randomness for effective rollout. This mechanism effectively bridges the gap between discrete token-level reasoning and continuous embedding-level exploration, enabling the model to explore a richer space that is inaccessible to standard token-level sampling. As illustrated in Figure~\ref{fig:pipeline}, we integrate this rollout paradigm into the state-of-the-art GRPO framework to propose \textbf{\Method}.

To validate the efficacy of our proposed framework, we conducted extensive evaluations across multiple challenging mathematical reasoning benchmarks. Experimental results demonstrate that our method consistently outperforms strong baselines in both Pass@16 and Pass@32 metrics across 1.5B and 7B model scales.

Our contributions are summarized as follows:

\begin{itemize}
  \item We propose a controllable embedding-level exploration method that injects semantically guided perturbations during generation.

  % \item We adapt this embedding-level exploration mechanism to the rollout stage of RL and present it as a general, plug-and-play module compatible with various rollout-based RL algorithms.

  % \item We implement this strategy within the GRPO framework and validate its effectiveness on challenging mathematical benchmarks, achieving superior performance compared to standard baselines.

  \item We integrate this mechanism into the GRPO framework to propose \textbf{\Method}, utilizing our Semantic Neighbor Mixing strategy.

  \item We validate the efficacy of our approach through extensive evaluations on challenging mathematical reasoning benchmarks, demonstrating that \Method consistently outperforms baselines across different model scales.
\end{itemize}

\section{Related Work}

\subsection{Reinforcement Learning for LLMs}

Reinforcement learning methods have become a cornerstone methodology for aligning Large Language Models (LLMs) and enhancing their complex reasoning capabilities~\cite{jaech2024openai,shao2024deepseekmath,yu2025dapo,zheng2025group,shrivastava2025sample}. Regardless of the specific optimization objective, a critical component shared across these paradigms is the rollout phase, where the policy must generate a diverse set of trajectories for accurate gradient estimation and advantage calculation. To facilitate this exploration and broaden the generation distribution, recent studies have introduced intrinsic exploration bonuses. COPO~\cite{bai2025online} modifies the objective function to incentivize the policy to visit under-explored or high-uncertainty states. Other approaches like min-p sampling~\cite{nguyen2024turning} intervene at decoding time by using a confidence-scaled cutoff. However, these strategies primarily rely on shallow token-level stochasticity. Consequently, they often struggle to induce semantic-level diversity without sacrificing coherence, limiting the model's ability to discover novel and high-reward reasoning paths.

A parallel line of work attempts to move exploration into the continuous latent space during RL rollouts. HRPO~\cite{yue2025hybrid} brings Soft Thinking~\cite{zhang2025soft} into GRPO by injecting mixed continuous representations into the context, but its randomness still fundamentally originates from discrete token sampling. SofT-GRPO~\cite{zheng2025soft} relies on Gumbel-reparameterized logits to enable soft-thinking-style policy optimization, where randomness again stems from token-level noise projected through a mixed embedding. More directly, \citet{butt2025soft} injects Gaussian noise on token embeddings or logits during rollout. As illustrated in Figure~\ref{fig:noise_vis}, such unconstrained noise can push representations off the semantic manifold and disrupt the rollout. In contrast, our \Method introduces embedding-level perturbations through a semantic neighbor set, reducing reliance on purely discrete token-level randomness while avoiding the semantic drift caused by unconstrained noise.

\subsection{Latent Reasoning}

Recently, due to the inefficiencies and limitations of discrete natural language reasoning, studies have begun to move beyond explicit Chain of Thought (CoT) to Latent Reasoning. In this paradigm, the intermediate reasoning process is carried by continuous hidden vectors instead of explicit text, thus decoupling thinking from language until the final answer is produced. Existing methods can be broadly categorized into two classes. The token-based class of approaches replaces or compresses CoT steps into latent states by training the model, e.g., the continuous thought feedback mechanism and curricular substitution of Coconut~\cite{hao2024training}, and the gist tokens with sparse attentional masks of LightThinker~\cite{zhang2025lightthinker} to reduce the long context cost and the use of key-value caches. However, mere latent adaptation is often subject to potential instabilities, such as semantic drift, and thus requires stabilization strategies such as SIM-COT~\cite{wei2025sim} with step-level supervision at training via an assistant decoder, and CODI~\cite{shen2025codi} with self-distillation alignment between explicit CoT and implicit CoT, in order to mitigate forgetting and improve robustness. Other approaches work on exploring compression and hybrid routing, where CoLaR~\cite{tan2025think} introduces controllable compression and reinforcement learning-based exploration for different latent trajectories, while SynAdapt~\cite{wang2025synadapt} utilizes synthesized continuous thought data and difficulty-aware routing.

\section{Method}

\subsection{Review of Standard Autoregressive Sampling}
\label{ssec:standard_sampling}

To facilitate subsequent derivations and descriptions, we first review the standard autoregressive generation process of language models.

Given an input prompt $q$, the policy model $\pi_\theta$ generates an output token sequence $o=(o_1,\dots,o_T)$ in an autoregressive manner, where $o_t\in\mathcal{V}$. At step $t$, conditioning on the prompt and the history of generated tokens $o_{<t}=(o_1,\dots,o_{t-1})$, the policy model computes the logits vector:
\begin{equation}
    z_t = \pi_\theta(q, o_{<t}) \in \mathbb{R}^{|\mathcal{V}|},
\end{equation}
where $z_t(v)$ represents the logit of token $v$.

Temperature sampling first applies temperature scaling (with $\tau>0$) to the logits vector:
\begin{equation}
    \label{eq:scaled_logits}
    \tilde{z}_t = \frac{z_t}{\tau},
\end{equation}
and obtains the token distribution at this step via the softmax function:
\begin{equation}
    p_t = \mathrm{softmax}(\tilde{z}_t).
\end{equation}

Standard sampling generates the output token randomly based on this distribution, which can be written as:
\begin{equation}
    \Pr(o_t=v\mid q,o_{<t})=p_t(v),\qquad \forall v\in\mathcal{V}.
\end{equation}

Subsequently, the embedding of the generated token $o_t$ is retrieved to serve as the input representation for the next time step $t+1$. Letting $e_{t+1}$ denote the input embedding at step $t+1$, we have:
\begin{equation}
    e_{t+1} = E[o_t],
\end{equation}
where $E \in \mathbb{R}^{|\mathcal{V}|\times d}$ denotes the token embedding matrix with vocabulary size $|\mathcal{V}|$ and dimension $d$, and $E[v]$ represents the continuous vector retrieved for a specific token $v\in\mathcal{V}$ via the lookup operation.

The process described above characterizes the fundamental mechanism of standard temperature sampling: at each time step, the model provides a vocabulary distribution based on the current context, samples a token from it, and feeds its corresponding embedding back as the input for the subsequent step to advance sequence generation.

\subsection{Semantic Neighbor Mixing}
\label{ssec:neighbor_mix}

Aiming to introduce perturbations directly at the embedding level during generation to expand the exploration scope of rollout trajectories, we construct an embedding-level sampling mechanism at each generation step: we identify the token most preferred by the current model as a semantic anchor token, retrieve its semantic neighbors in the embedding space, and allocate weights within this set of neighbor tokens, using the current step's logits to obtain a continuous mixed embedding as the input for the next step.

Specifically, given the temperature-scaled logits $\tilde{z}_t$ obtained from Equation~\ref{eq:scaled_logits}, we first determine the semantic anchor token:
\begin{equation}
    v_t^* = \arg\max_{v\in\mathcal{V}} \tilde{z}_t(v).
\end{equation}
By using the argmax, we ensure that the exploration center aligns perfectly with the model's optimal reasoning path, avoiding the instability caused by sampling low-probability tokens as anchors.

Subsequently, we retrieve a neighbor set $\mathcal{C}_t$ of size $k$ surrounding the anchor in the embedding matrix $E$, which includes the anchor itself:
\begin{equation}
    \mathcal{C}_t = \{v_t^{(1)},\dots,v_t^{(k)}\},\qquad v_t^{(1)}=v_t^*.
\end{equation}

Specifically, we employ cosine similarity in the embedding space to measure the semantic relevance between tokens:
\begin{equation}
    s(u,v) = \frac{E[u] \cdot E[v]}{\|E[u]\|_2 \|E[v]\|_2},
\end{equation}
and accordingly select the $k-1$ tokens with the highest similarity to the anchor from $\mathcal{V}\setminus\{v_t^*\}$ to form $\mathcal{C}_t$ together with the anchor token.

After obtaining the candidate set $\mathcal{C}_t$, we normalize the temperature-scaled logits solely over this set to obtain mixing weights. For any candidate token $c\in\mathcal{C}_t$, let:
\begin{equation}
    \alpha_t(c)=\frac{\exp(\tilde{z}_t(c))}{\displaystyle\sum_{u\in\mathcal{C}_t}\exp(\tilde{z}_t(u))}.
\end{equation}

Finally, we construct the mixed embedding:
\begin{equation}
    \tilde{e}_{t+1}=\sum_{c\in\mathcal{C}_t}\alpha_t(c)E[c],
\end{equation}
and use $\tilde{e}_{t+1}$ as the input representation for the next generation step. This process introduces continuous perturbations within the local semantic neighborhood of the anchor token. The direction and intensity of these perturbations are adaptively modulated by the logits of the current step, thereby expanding the exploration range while maintaining consistency with the model's current semantic preferences as much as possible.

\subsection{Adaptation to the GRPO}

We present \textbf{\Method} by applying the aforementioned embedding-level sampling to the rollout phase of GRPO (Group Relative Policy Optimization)~\cite{shao2024deepseekmath} while preserving GRPO's advantage estimation and optimization objectives. In each update, for every input $q$, GRPO samples a group of outputs $\{o_i\}_{i=1}^{G}$ of size $G$ from the old policy $\pi_{\theta_{\mathrm{old}}}$ and performs updates based on group-relative advantages.

To enhance the diversity and exploration efficiency of the generated trajectories, we introduce a gating mechanism during the rollout process: with a fixed probability $\rho$ as mixing rate, we replace the standard token-level sampling with the embedding-level mixing described in Section~\ref{ssec:neighbor_mix}, while continuing to use standard temperature sampling for the remaining time steps. This stochastic gating mechanism effectively preserves the semantic stability of the generation by keeping the majority of steps to the discrete tokens. Simultaneously, it introduces necessary intermittency in exploration, which prevents the policy from becoming overly reliant on continuous perturbations, thereby mitigating potential risks of training instability and semantic drift.

Specifically, for the $t$-th step of the $i$-th sequence, let $m_{i,t}\sim\mathrm{Bernoulli}(\rho)$ serve as a mixing mask. This mask dictates whether the model follows the standard trajectory or activates the Semantic Neighbor Mixing mechanism. The input representation $e_{i,t+1}$ for the subsequent step is then determined as follows:
\begin{equation}
    e_{i,t+1} = 
    \begin{cases} 
    \displaystyle \sum_{c\in\mathcal{C}_{i,t}}\alpha_{i,t}(c)E[c], & \text{if } m_{i,t}=1, \\
    E[o_{i,t}], & \text{else},
    \end{cases}
\end{equation}
where in the case of $m_{i,t}=1$, the model activates Semantic Neighbor Mixing introduced in Section~\ref{ssec:neighbor_mix}. Otherwise (i.e., $m_{i,t}=0$), the model follows standard sampling, where the token $o_{i,t}$ is sampled as described in Section~\ref{ssec:standard_sampling} and the input is given by its embedding $E[o_{i,t}]$.

In terms of implementation, to ensure the rollout trajectory is reproducible during the training phase, we record the candidate set and the associated weights $\{(\mathcal{C}_{i,t}, \alpha_{i,t})\}$ in the rollout buffer. During the training phase, the input representation $e_{i,t}$ is reconstructed via table lookup based on these records, ensuring that the calculation of probabilities and losses remains consistent with the rollout generation.

Since the final answer verification and the reward calculation typically require discrete text input, we use the anchor token $v_t^*$ as the textual realization for steps where latent mixing occurred. Let $\tilde{o}_i$ denote the discrete token sequence where mixed steps are replaced by their anchors. The advantages are computed based on the reward of this discrete sequence:
\begin{equation}
    \hat{A}_{i} = \frac{r(\tilde{o}_i)-\mathrm{mean}(r)}{\mathrm{std}(r)}.
\end{equation}
This design allows us to leverage standard reward while performing exploration in the continuous embedding space.

The optimization follows the standard GRPO objective, which incorporates a PPO-style clipped surrogate and a KL regularization term:
\begin{equation}
\begin{split}
    J_{\mathrm{GRPO}}(\theta) = \mathbb{E}\Bigg[ \frac{1}{G}\sum_{i=1}^{G}\frac{1}{|o_i|}\sum_{t=1}^{|o_i|} \bigg( & \\
    \min\Big(r_{i,t}(\theta)\hat{A}_{i},\ \mathrm{clip}\big(r_{i,t}(\theta),&1-\epsilon,1+\epsilon\big)\hat{A}_{i}\Big) \\
    -\beta D_{\mathrm{KL}}\left(\pi_\theta(\cdot|h_{i,t})\ ||\ \pi_{\mathrm{ref}}(\cdot|h_{i,t})\right) & \bigg) \Bigg],
\end{split}
\end{equation}
where $r_{i,t}(\theta) = \frac{\pi_\theta(o_{i,t}|h_{i,t})}{\pi_{\theta_{\mathrm{old}}}(o_{i,t}|h_{i,t})}$ denotes the importance ratio conditioned on the reconstructed history $h_{i,t}$, and $\beta$ is the KL penalty coefficient.

\section{Experiments}

\begin{table*}[ht]
\small
\centering
\setlength{\tabcolsep}{5pt}
\begin{tabular}{@{}lcccccccccccc@{}}
\toprule
\multicolumn{1}{l|}{} & \multicolumn{3}{c|}{\textbf{AIME25}} & \multicolumn{3}{c|}{\textbf{AMC23}} & \multicolumn{3}{c|}{\textbf{MATH500}} & \multicolumn{3}{c}{\textbf{Average}} \\
\multicolumn{1}{l|}{} & @1 & @16 & \multicolumn{1}{c|}{@32} & @1 & @16 & \multicolumn{1}{c|}{@32} & @1 & @16 & \multicolumn{1}{c|}{@32} & @1 & @16 & @32 \\ \midrule
\multicolumn{13}{c}{\textit{\textbf{DeepSeek-R1-Distill-Qwen-1.5B}}} \\ \midrule
\multicolumn{1}{l|}{Base Model} & 20.83 & 38.05 & \multicolumn{1}{c|}{41.19} & 64.14 & 91.73 & \multicolumn{1}{c|}{92.37} & 74.88 & 89.38 & \multicolumn{1}{c|}{90.29} & 53.28 & 73.05 & 74.62 \\
\multicolumn{1}{l|}{+Soft Thinking} & 20.73 & 39.90 & \multicolumn{1}{c|}{44.56} & \textbf{64.77} & {\ul 94.03} & \multicolumn{1}{c|}{96.09} & 75.12 & {\ul 89.77} & \multicolumn{1}{c|}{{\ul 90.90}} & 53.54 & 74.56 & 77.18 \\
\multicolumn{1}{l|}{+GRPO} & 21.35 & 43.61 & \multicolumn{1}{c|}{{\ul 47.31}} & 63.91 & 93.42 & \multicolumn{1}{c|}{94.60} & \textbf{75.61} & 89.43 & \multicolumn{1}{c|}{90.32} & 53.62 & 75.49 & 77.41 \\
\multicolumn{1}{l|}{+GRPO+Soft Thinking} & {\ul 22.50} & 43.94 & \multicolumn{1}{c|}{45.94} & 63.75 & 93.91 & \multicolumn{1}{c|}{95.96} & {\ul 75.53} & 89.60 & \multicolumn{1}{c|}{90.70} & {\ul 53.93} & {\ul 75.82} & 77.53 \\
\multicolumn{1}{l|}{+STHT} & 20.42 & 42.66 & \multicolumn{1}{c|}{46.73} & 62.03 & 93.75 & \multicolumn{1}{c|}{\textbf{96.88}} & 74.81 & 89.43 & \multicolumn{1}{c|}{90.55} & 52.42 & 75.28 & {\ul 78.05} \\
\multicolumn{1}{l|}{+\Method} & \textbf{23.33} & \textbf{46.48} & \multicolumn{1}{c|}{\textbf{50.28}} & {\ul 64.14} & \textbf{94.07} & \multicolumn{1}{c|}{{\ul 96.18}} & 74.87 & \textbf{89.92} & \multicolumn{1}{c|}{\textbf{91.05}} & \textbf{54.11} & \textbf{76.82} & \textbf{79.17} \\ \midrule
\multicolumn{13}{c}{\textit{\textbf{DeepSeek-R1-Distill-Qwen-7B}}} \\ \midrule
\multicolumn{1}{l|}{Base Model} & 33.23 & 54.73 & \multicolumn{1}{c|}{57.38} & 79.92 & 93.50 & \multicolumn{1}{c|}{94.16} & 84.24 & 91.66 & \multicolumn{1}{c|}{92.14} & 65.80 & 79.96 & 81.23 \\
\multicolumn{1}{l|}{+Soft Thinking} & 32.60 & 54.17 & \multicolumn{1}{c|}{57.12} & 80.08 & 93.96 & \multicolumn{1}{c|}{94.70} & 84.01 & 92.08 & \multicolumn{1}{c|}{92.59} & 65.56 & 80.07 & 81.47 \\
\multicolumn{1}{l|}{+GRPO} & {\ul 34.27} & 56.13 & \multicolumn{1}{c|}{58.32} & {\ul 81.88} & 94.28 & \multicolumn{1}{c|}{94.87} & \textbf{85.40} & 92.21 & \multicolumn{1}{c|}{92.63} & {\ul 67.18} & 80.87 & 81.94 \\
\multicolumn{1}{l|}{+GRPO+Soft Thinking} & \textbf{34.58} & {\ul 56.18} & \multicolumn{1}{c|}{{\ul 58.91}} & \textbf{82.81} & 94.16 & \multicolumn{1}{c|}{94.70} & {\ul 85.39} & {\ul 92.44} & \multicolumn{1}{c|}{{\ul 93.03}} & \textbf{67.59} & 80.93 & 82.21 \\
\multicolumn{1}{l|}{+STHT} & 33.75 & 55.76 & \multicolumn{1}{c|}{58.40} & 80.23 & {\ul 95.43} & \multicolumn{1}{c|}{{\ul 96.49}} & 83.95 & 92.08 & \multicolumn{1}{c|}{92.71} & 65.98 & {\ul 81.09} & {\ul 82.53} \\
\multicolumn{1}{l|}{+\Method} & {\ul 34.27} & \textbf{58.00} & \multicolumn{1}{c|}{\textbf{61.18}} & 80.70 & \textbf{96.92} & \multicolumn{1}{c|}{\textbf{98.13}} & 84.94 & \textbf{92.77} & \multicolumn{1}{c|}{\textbf{93.28}} & 66.64 & \textbf{82.56} & \textbf{84.20} \\ \midrule
\multicolumn{13}{c}{\textit{\textbf{Llama-3.2-1B}}} \\ \midrule
\multicolumn{1}{l|}{Base Model} & 0.10 & 1.31 & \multicolumn{1}{c|}{2.15} & 11.33 & 50.28 & \multicolumn{1}{c|}{62.44} & 15.64 & 52.84 & \multicolumn{1}{c|}{60.23} & 9.03 & 34.81 & 41.61 \\
\multicolumn{1}{l|}{+GRPO} & {\ul 0.21} & {\ul 2.72} & \multicolumn{1}{c|}{{\ul 4.39}} & \textbf{15.16} & \textbf{54.29} & \multicolumn{1}{c|}{{\ul 63.86}} & \textbf{31.97} & 61.93 & \multicolumn{1}{c|}{66.07} & \textbf{15.78} & {\ul 39.65} & {\ul 44.77} \\
\multicolumn{1}{l|}{+STHT} & {\ul 0.21} & 2.64 & \multicolumn{1}{c|}{4.18} & 11.25 & 49.62 & \multicolumn{1}{c|}{61.19} & 27.03 & {\ul 62.04} & \multicolumn{1}{c|}{{\ul 67.15}} & 12.83 & 38.10 & 44.17 \\
\multicolumn{1}{l|}{+\Method} & \textbf{0.31} & \textbf{3.49} & \multicolumn{1}{c|}{\textbf{5.09}} & {\ul 11.95} & {\ul 54.09} & \multicolumn{1}{c|}{\textbf{65.88}} & {\ul 27.54} & \textbf{62.44} & \multicolumn{1}{c|}{\textbf{68.04}} & {\ul 13.27} & \textbf{40.01} & \textbf{46.34} \\ \midrule
\multicolumn{13}{c}{\textit{\textbf{Qwen3-1.7B-Base}}} \\ \midrule
\multicolumn{1}{l|}{Base Model} & 2.81 & 19.15 & \multicolumn{1}{c|}{23.18} & 30.78 & 69.90 & \multicolumn{1}{c|}{76.40} & 50.37 & {\ul 80.84} & \multicolumn{1}{c|}{{\ul 83.71}} & 27.99 & 56.63 & 61.09 \\
\multicolumn{1}{l|}{+GRPO} & \textbf{5.52} & {\ul 20.03} & \multicolumn{1}{c|}{23.47} & \textbf{43.28} & \textbf{74.48} & \multicolumn{1}{c|}{\textbf{79.35}} & \textbf{64.76} & \textbf{84.73} & \multicolumn{1}{c|}{\textbf{86.22}} & \textbf{37.85} & \textbf{59.75} & {\ul 63.01} \\
\multicolumn{1}{l|}{+STHT} & 2.40 & 19.52 & \multicolumn{1}{c|}{{\ul 25.78}} & 30.00 & 70.57 & \multicolumn{1}{c|}{76.53} & 51.29 & 80.71 & \multicolumn{1}{c|}{83.51} & 27.89 & 56.93 & 61.94 \\
\multicolumn{1}{l|}{+\Method} & {\ul 3.23} & \textbf{22.87} & \multicolumn{1}{c|}{\textbf{28.47}} & {\ul 30.86} & {\ul 71.49} & \multicolumn{1}{c|}{{\ul 78.25}} & {\ul 52.86} & 80.70 & \multicolumn{1}{c|}{83.66} & {\ul 28.98} & {\ul 58.36} & \textbf{63.46} \\ \bottomrule
\end{tabular}
\caption{\textbf{Experimental results of baselines and our method on mathematical reasoning benchmarks.} We evaluate DeepSeek-R1-Distill-Qwen, Llama-3.2-1B, and Qwen3-1.7B-Base backbones. STHT denotes the Soft Tokens, Hard Truths baseline~\cite{butt2025soft}. The columns @1, @16, and @32 denote the Mean@32 (average accuracy), Pass@16, and Pass@32 scores, respectively. All values are reported in percentages. \textbf{Bold} values indicate the best performance, while \underline{underlined} values indicate the second-best performance in each column for the respective model size.}
\label{tab:math}
\end{table*}

We evaluate \Method on both reasoning-distilled and non-distilled backbones. DeepSeek-R1-Distill-Qwen-1.5B and 7B serve as the primary reasoning-distilled models, allowing us to assess performance on strong math-oriented policies at different scales. Llama-3.2-1B and Qwen3-1.7B-Base are included as non-distilled base models to test whether the gains depend on reasoning distillation or a particular backbone family.

\subsection{Implementation Details}
\label{ssec:implementation_details}

\paragraph{Baselines.}
We benchmark \Method against several representative baselines. The untrained \textbf{Base Model} and standard \textbf{GRPO}~\cite{shao2024deepseekmath} serve as foundational performance references. \textbf{Soft Thinking}~\cite{zhang2025soft} is a training-free inference-time strategy that replaces the sampled discrete token with a probability-weighted sum over the entire vocabulary embedding at each decoding step, without any parameter optimization. We additionally combine it with GRPO-trained policies (\textbf{GRPO+Soft Thinking}) to examine whether its gains stack with RL. We also include \textbf{Soft Tokens, Hard Truths (STHT)}~\cite{butt2025soft}, which adds Gaussian perturbations to token embeddings during rollout and provides a direct embedding-space perturbation baseline.

\paragraph{Training Settings.}
We implement our training pipeline using the \texttt{verl} framework, extending the codebase with custom modules to support our proposed method. For all variants, we use the training dataset released by DeepScaleR~\cite{luo2025deepscaler}. To ensure training efficiency and stability, we filter out samples with input prompts exceeding 4,096 tokens. This threshold is selected to balance the inclusion of complex problem statements with the memory constraints of the attention mechanism. Furthermore, the maximum generation length is set to 8,192 tokens. This context window is critical for accommodating the lengthy chain-of-thought rationales required for advanced mathematical derivations, ensuring that the model's reasoning process is not prematurely truncated before reaching the final answer. All runs are trained with a learning rate of 1e-6 and a global batch size of 64, distributed across 8 NVIDIA H20-3e GPUs. Unless otherwise specified, we set the mixing rate $\rho$ to 0.1 and $k$ to 3 in our proposed method. Each experiment is trained for one epoch. We monitor validation performance on AIME24 during training and select the best checkpoint as the final model for subsequent evaluation. For a complete list of hyperparameter settings and prompts, please refer to Appendix~\ref{app:task_prompt} and~\ref{app:implementation_details}.

\paragraph{Evaluation Settings.}
We report in-domain math reasoning performance on AMC23, AIME25, and MATH500~\cite{hendrycks2021measuring}. To assess out-of-distribution generalization, we further evaluate on GPQA-Diamond, a subset of GPQA~\cite{rein2024gpqa}. Following the recommended settings in DeepSeek-R1~\cite{guo2025deepseek}, we set the sampling temperature to 0.6 and top-p to 0.95. We evaluate performance using Mean@32, Pass@16, and Pass@32.

\subsection{Evaluation on Math Benchmarks}

As shown in Table~\ref{tab:math}, across the two DeepSeek-R1-Distill-Qwen model sizes, \Method achieves the best average Pass@16 and Pass@32 and improves most per-benchmark Pass@k metrics. For the 1.5B model, average Pass@32 increases from 74.62 for the base model and 77.41 for GRPO to 79.17 with \Method. On the challenging AIME25 benchmark, Pass@32 rises from 41.19 for the base model and 47.31 for GRPO to 50.28 with \Method. Soft Thinking alone does not consistently improve over GRPO-trained policies, while our mean@32 results remain comparable to other methods. Compared with STHT, which injects unconstrained Gaussian noise, \Method improves average Pass@32 from 78.05 to 79.17 at 1.5B and from 82.53 to 84.20 at 7B, although STHT remains competitive on AMC23. This trend supports our motivation that embedding-level exploration benefits from semantically grounded perturbations rather than unstructured noise.

For the additional non-distilled backbones, the gains are most visible in aggregate Pass@k and on the harder AIME25 benchmark. On Llama-3.2-1B, \Method attains the highest average Pass@16 and Pass@32, and also gives the best Pass@32 on all three math benchmarks. On the stronger Qwen3-1.7B-Base, GRPO remains stronger on AMC23 and MATH500, but \Method improves AIME25 Pass@32 over GRPO by 5.00 points and achieves the highest overall average Pass@32. Since AMC23 and MATH500 already have higher baseline pass rates, this pattern suggests that semantic neighbor mixing is most helpful when the benchmark leaves more room for exploration. The consistent improvements on both Llama and Qwen3 demonstrate that the gain is not specific to a single tokenizer or model family. These results indicate that the benefit of embedding-level neighbor mixing is not tied to reasoning-distilled policies.

\subsection{Evaluation on Out-of-Distribution (OOD) Benchmarks}

\begin{figure*}[ht]
  \centering
  \includegraphics[width=\linewidth]{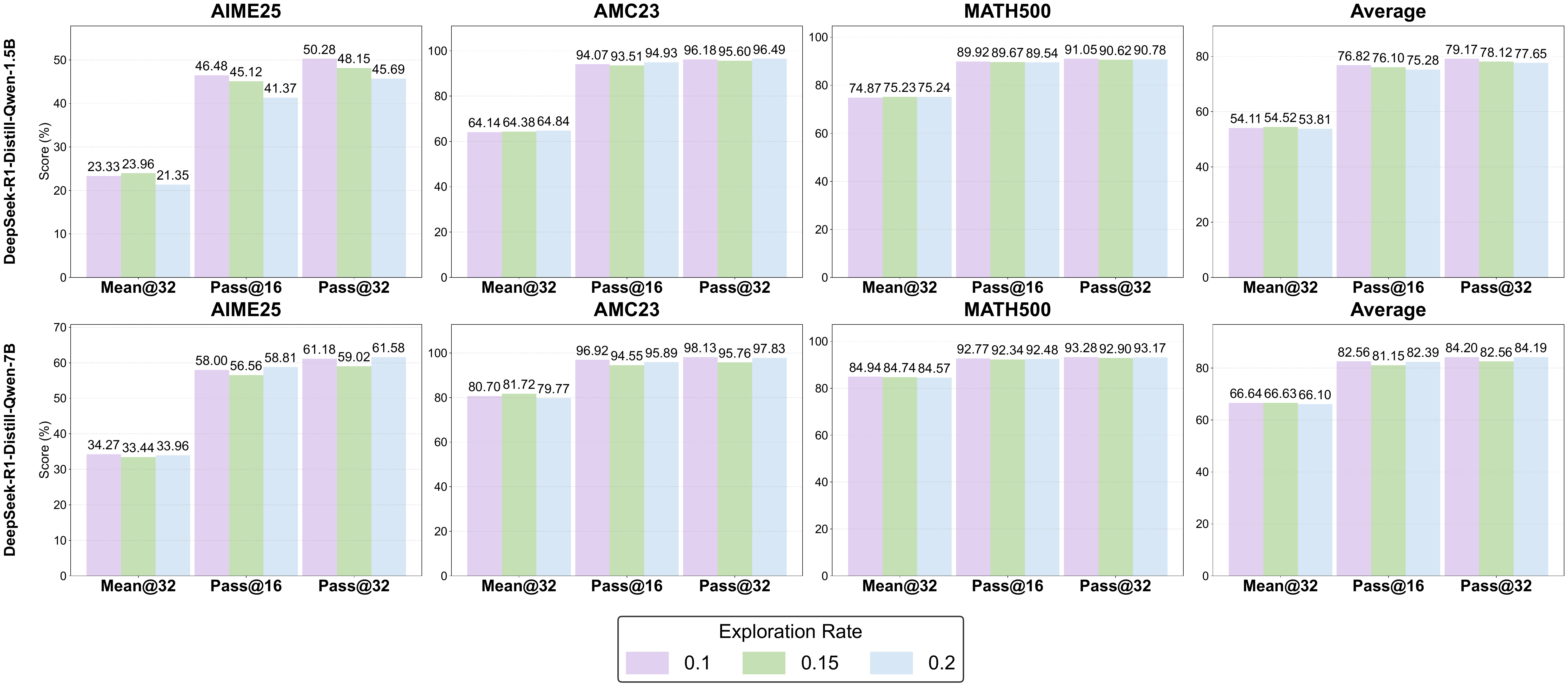}
  \caption{\textbf{Performance evaluation of DeepSeek-R1-Distill-Qwen models across math benchmarks with varying mixing rates.} The charts illustrate the Mean@32, Pass@16, and Pass@32 metrics, with the last column showing the average performance across all datasets.}
  \label{fig:exploration_rate}
\end{figure*}

\begin{table}[t]
\small
\centering
\setlength{\tabcolsep}{4pt}
\begin{tabular}{@{}lccc@{}}
\toprule
 & Mean@32 & Pass@16 & Pass@32 \\ \midrule
\multicolumn{4}{c}{\textit{\textbf{DeepSeek-R1-Distill-Qwen-1.5B}}} \\ \midrule
\multicolumn{1}{l|}{Base Model} & 32.23 & 85.51 & 90.79 \\
\multicolumn{1}{l|}{+Soft Thinking} & 32.86 & 86.31 & 91.73 \\
\multicolumn{1}{l|}{+GRPO} & 32.24 & 86.25 & 91.95 \\
\multicolumn{1}{l|}{+GRPO+Soft Thinking} & 33.19 & 86.89 & 91.92 \\
\multicolumn{1}{l|}{+\Method} & \textbf{33.68} & \textbf{87.58} & \textbf{92.87} \\ \midrule
\multicolumn{4}{c}{\textit{\textbf{DeepSeek-R1-Distill-Qwen-7B}}} \\ \midrule
\multicolumn{1}{l|}{Base Model} & 46.59 & 86.55 & 90.73 \\
\multicolumn{1}{l|}{+Soft Thinking} & 45.99 & 86.96 & 91.28 \\
\multicolumn{1}{l|}{+GRPO} & 47.00 & 87.15 & 91.42 \\
\multicolumn{1}{l|}{+GRPO+Soft Thinking} & 46.78 & 87.57 & 92.29 \\
\multicolumn{1}{l|}{+\Method} & \textbf{47.16} & \textbf{88.77} & \textbf{92.80} \\ \bottomrule
\end{tabular}
\caption{\textbf{Performance comparison on the GPQA-Diamond benchmark for out-of-distribution (OOD) evaluation.} We report results for DeepSeek-R1-Distill-Qwen models (1.5B and 7B). \textbf{Bold} values indicate the best performance in each column.}
\label{tab:gpqa}
\end{table}

To verify that \Method improves mathematical reasoning without overfitting to the training distribution, we evaluate its out-of-distribution (OOD) performance on the GPQA-Diamond benchmark. As shown in Table~\ref{tab:gpqa}, the results demonstrate that our approach not only excels in mathematical domains but also generalizes effectively to complex scientific reasoning tasks, achieving consistent gains in both stability (mean@32) and peak performance (pass@32). For the 1.5B model, our method consistently outperforms the base model and other baselines. Specifically, we achieve a pass@32 score of 92.87, representing a clear improvement over the base model (90.79) and surpassing the combination of GRPO and Soft Thinking (91.92). Similar trends are observed with the 7B model, where our method attains the highest performance across all metrics. These results confirm that our approach not only strengthens specific mathematical reasoning skills but also preserves and enhances the model's general scientific reasoning capabilities.

\subsection{Transfer to GSPO}
\label{ssec:gspo_transfer}

To examine whether Semantic Neighbor Mixing can transfer beyond the GRPO training recipe, we further apply the same module to GSPO, denoted \textbf{N-GSPO}, on DeepSeek-R1-Distill-Qwen-1.5B. The results show that N-GSPO improves average Pass@32 from 77.34 to 79.04 over GSPO, with a $+7.66$ gain on AIME25. This suggests that the proposed perturbation mechanism is not tightly coupled to the specific advantage normalization used by GRPO. Full per-benchmark numbers are reported in Appendix~\ref{app:ngspo}.

\subsection{Analysis of \Method}

To provide a deeper understanding of the mechanisms driving the performance improvements, we conduct a series of ablation studies and sensitivity analyses. In this section, we systematically isolate key components to verify the stability and effectiveness of \Method under varying conditions. Specifically, we examine the sensitivity to mixing rates, the impact of different mixing strategies, and the choice of distance measurements.

\subsubsection{Impact of Mixing Rates}

To investigate the robustness of our method against hyperparameter variations, we conducted an ablation study on different mixing rates. As illustrated in Figure~\ref{fig:exploration_rate}, the overall performance remains relatively stable across varying mixing rates for both the 1.5B and 7B models. The average results show only marginal fluctuations, indicating that our approach is generally insensitive to minor adjustments in mixing rate. Complete tabulated results can be found in Appendix~\ref{app:full_ablation_results}.

However, a notable exception is observed in the AIME25 benchmark with the smaller 1.5B model. When the mixing rate is increased to 0.2, the Pass@32 metric drops significantly. We attribute this phenomenon to the trade-off between diversity and stability in the reasoning process. While an appropriate level of embedding-level exploration encourages diverse reasoning paths, excessive embedding-level exploration can disrupt the semantic stability of the generated rollouts.

\subsubsection{Impact of Mixing Mechanisms}

\begin{table}[ht]
\small
\centering
\setlength{\tabcolsep}{3pt}
\begin{tabular}{@{}l|ccc@{}}
\toprule
 & Mean@32 & Pass@16 & Pass@32 \\ \midrule
DS-R1-Distill-Qwen-1.5B & 53.28 & 73.05 & 74.62 \\
+Gumbel Soft-Thinking & 53.87 & 74.58 & 76.89 \\
+\Method w/o rate & 53.56 & 75.16 & 77.32 \\
+\Method & \textbf{54.11} & \textbf{76.82} & \textbf{79.17} \\ \bottomrule
\end{tabular}
\caption{\textbf{Ablation study on mixing strategies.} We report the average performance across the mathematical benchmarks mentioned in Section~\ref{ssec:implementation_details} using the DeepSeek-R1-Distill-Qwen-1.5B model. "+Gumbel Soft-Thinking" replaces our mixing mechanism with Gumbel noise-based top-k selection. "+\Method w/o rate" applies the method to all tokens, removing the mixing rate constraint.}
\label{tab:mix_strategy}
\end{table}

To investigate the effectiveness of our proposed mixing mechanism and the necessity of the mixing rate, we conduct an ablation study using the 1.5B model. Table~\ref{tab:mix_strategy} summarizes the average results across the mathematical benchmarks utilized in our main experiments. Detailed performance breakdowns for each individual benchmark are provided in Appendix~\ref{app:full_ablation_results}.

We compare our approach with Gumbel Soft-Thinking proposed in \citet{zheng2025soft}, which injects Gumbel noise into logits for weighted mixing. This method serves as a representative of direct noise injection into logits or embeddings, a category of strategies we have discussed in Section~\ref{sec:introduction}. While this baseline improves over the vanilla model, our method achieves superior average performance, suggesting that our strategy provides more effective exploration guidance than native noise in embeddings.

Additionally, we evaluate \Method w/o rate, where the mixing operation is applied to every token. The observed drop in performance indicates that applying our mixing mechanism to all tokens introduces excessive noise, identifying the mixing rate as a critical filter for maintaining rollout stability.

\subsubsection{Impact of Distance Measurements}

\begin{table}[t]
\small
\centering
\setlength{\tabcolsep}{8pt}
\begin{tabular}{@{}l|ccc@{}}
\toprule
 & Mean@32 & Pass@16 & Pass@32 \\ \midrule
Cosine & \textbf{54.11} & \textbf{76.82} & \textbf{79.17} \\
L2 & 53.77 & 75.26 & 77.68 \\
L1 & 53.07 & 72.98 & 75.16 \\ \bottomrule
\end{tabular}
\caption{\textbf{Performance comparison of different distance metrics for neighbor selection.} We evaluate Cosine distance, L2 distance, and L1 distance on the DeepSeek-R1-Distill-Qwen-1.5B model.}
\label{tab:distance_measurement}
\end{table}

The choice of distance measurement is crucial for quantifying the relations between representations in our mixing mechanism. We compare three widely used metrics: Cosine distance, L2 (Euclidean) distance, and L1 (Manhattan) distance.

As shown in Table~\ref{tab:distance_measurement}, which reports the average performance of the 1.5B model across the aforementioned mathematical benchmarks, Cosine distance significantly outperforms both L1 and L2 metrics, achieving the highest scores across all stability and performance benchmarks. A complete listing of performance metrics for each distance variant can be found in Appendix~\ref{app:full_ablation_results}.

This performance gap can be attributed to the geometric properties of the high-dimensional embedding space used by LLMs. Semantic information in these representations is primarily encoded in the direction of the embedding vectors rather than their magnitude~\cite{schakel2015measuring,kozlowski2025semantic}. Cosine distance effectively captures this directional alignment, whereas L2 distance is sensitive to vector norms, which may introduce irrelevant variance and degrade the quality of the embedding-level exploration.

\subsection{Impact of Semantic Neighbor Mixing During Inference}

\begin{table}[ht]
\small
\centering
\setlength{\tabcolsep}{3pt}
\begin{tabular}{@{}lccc@{}}
\toprule
\multicolumn{1}{c|}{Inference Strategy} & Mean@32 & Pass@16 & Pass@32 \\ \midrule
\multicolumn{4}{c}{\textit{\textbf{\Method trained DeepSeek-R1-Distill-Qwen-1.5B}}} \\ \midrule
\multicolumn{1}{l|}{Standard} & \textbf{54.11} & \textbf{76.82} & \textbf{79.17} \\
\multicolumn{1}{l|}{Semantic Neighbor Mixing} & 53.19 & 74.67 & 77.05 \\ \midrule
\multicolumn{4}{c}{\textit{\textbf{\Method trained DeepSeek-R1-Distill-Qwen-7B}}} \\ \midrule
\multicolumn{1}{l|}{Standard} & \textbf{66.64} & \textbf{82.56} & \textbf{84.20} \\
\multicolumn{1}{l|}{Semantic Neighbor Mixing} & 65.69 & 80.00 & 81.45 \\ \bottomrule
\end{tabular}
\caption{\textbf{Performance comparison of different inference strategies.} We evaluate the impact of applying Semantic Neighbor Mixing at the inference phase, compared to standard temperature sampling.}
\label{tab:inference_strategy}
\end{table}

While \Method utilizes Semantic Neighbor Mixing during the rollout phase of training to enhance exploration, we observe that applying this same mixing strategy during the inference phase leads to a degradation in performance. As shown in Table~\ref{tab:inference_strategy}, which reports the average performance across all evaluated mathematical benchmarks, enabling Semantic Neighbor Mixing at test time results in lower accuracy compared to standard temperature sampling. Comprehensive results for each specific dataset are detailed in Appendix~\ref{app:full_ablation_results}.

We attribute this phenomenon to the distinct objectives of the rollout and inference phases. During rollout, the mixing strategy acts as an exploration mechanism. When it introduces noise to explore more trajectories, the GRPO framework effectively filters new low-value trajectories. However, during inference, the goal shifts to maximizing response accuracy, where such a filtering mechanism is absent. Deviations caused by mixing may undermine logical coherence, consequently degrading the quality of the final solution. Therefore, we disable mixing to avoid introducing unnecessary noise that could disrupt the rigorous mathematical reasoning.

\section{Conclusion}

We have presented a novel embedding-level exploration strategy that enhances the efficiency of reinforcement learning for LLMs. By shifting the paradigm from discrete token-level randomness to continuous embedding-level perturbations, our method introduces semantically grounded noise that encourages the model to traverse diverse reasoning paths without sacrificing semantic stability. We implemented this mechanism via Semantic Neighbor Mixing and integrated it seamlessly into the GRPO framework, resulting in our proposed approach, \Method. Extensive experiments across multiple model scales demonstrate that our approach consistently outperforms baselines on challenging mathematical reasoning benchmarks, while also exhibiting superior generalization on out-of-distribution scientific tasks. These findings suggest that exploring reasoning dynamics in the embedding space offers a promising avenue for advancing LLM reasoning capabilities.
% placeholder placeholder placeholder placeholder placeholder placeholder placeholder placeholder placeholder placeholder placeholder placeholder placeholder placeholder placeholder placeholder placeholder placeholder placeholder placeholder placeholder placeholder placeholder 

\section*{Limitations}

The proposed Semantic Neighbor Mixing mechanism inevitably introduces additional computational overhead during the rollout phase. Unlike standard sampling which operates directly on computed logits, our approach requires retrieving nearest neighbors and calculating weighted embeddings at generation steps, which may increase inference latency. Furthermore, our experimental validation has been primarily concentrated on mathematical and scientific reasoning benchmarks. While the results indicate strong performance in these areas, we have not yet verified the effectiveness of this embedding-level exploration strategy on code generation tasks. Given that coding problems often require distinct structural logic and strict syntax constraints compared to natural language reasoning, extending our method to the programming domain remains an important subject for future investigation.

\section*{Acknowledgments}

This work was supported by the National Natural Science Foundation of China (62441617) and "Pioneer" and "Leading Goose" R\&D Program of Zhejiang (No.2024C01142). This work was also supported by Ant Group through CAAI-Ant Research Fund and the Earth System Big Data Platform of the School of Earth Sciences, Zhejiang University.

\bibliography{custom}

\clearpage
\appendix

\section{Task Prompt}
\label{app:task_prompt}

\begin{promptbox}[title=Prompt Example for Math Task]
\texttt{<|User|>} \\
\textcolor[HTML]{B07AD8}{An isosceles trapezoid has an inscribed circle tangent to each of its four sides. The radius of the circle is $3$, and the area of the trapezoid is $72$. Let the parallel sides of the trapezoid have lengths $r$ and $s$, with $r \neq s$. Find $r^2+s^2$} Let's think step by step and output the final answer within \textbackslash boxed{}. \\
\texttt{<|Assistant|>}<think>
\end{promptbox}

\begin{promptbox}[title=Prompt Example for GPQA-Diamond]
\texttt{<|User|>} \\
{
\color[HTML]{B07AD8}
Very large number of neutrinos produced by the Sun reach the Earth (very large flux of neutrinos, defined as the number of neutrinos per cm\textsuperscript{2}, per second). \\ \\
Let us assume that, hypothetically, the pp-III branch suddenly stopped in the core of the Sun about 8 and a half minutes ago, while all other reactions remained as they were. \\ \\
What would be the approximate ratio of the flux between two bands of neutrino energies of 700-800 KeV (band 1) and 800-900 keV (band 2). \\ \\
Flux (band 1) / flux (band 2) is:  \\ \\
(Note: we are talking about stopping the pp-III branch, not pp-II, pp-I or any other. It's not a typo or something like that.) \\
(Note 2: solar neutrino flavor changes happen, but do not play a role here.)
} \\ \\
{
\color[HTML]{F4B183}
A. 0.1 ($10^{-1}$). \\
B. 0.01 ($10^{-2}$). \\
C. 1. \\
D. 10.
} \\
Let's think step by step and output the final answer in the last line as: \\
Answer: <A/B/C/D> \\
\texttt{<|Assistant|>}<think>
\end{promptbox}

This section presents the specific prompt templates used during both the training and evaluation phases. To elicit the model's reasoning capabilities and standardize the output format, we employ an explicit Chain-of-Thought (CoT) instruction format. All input sequences consist of the user query followed by specific formatting constraints.

As illustrated above, for mathematical reasoning tasks (AMC23, AIME24, AIME25, and MATH500), the model is instructed to enclose the final result within \textbackslash boxed{}. For multiple-choice tasks such as GPQA, a specific answer format is enforced. Consistent with the model's default settings, we directly append the <think> token to the prompt. This forces the model to initiate the reasoning process.

\section{Implementation Details}
\label{app:implementation_details}

\subsection{Codebase and Framework}

Our implementation is developed based on the \texttt{verl}\footnote{\url{https://github.com/volcengine/verl}} framework (version 0.5.0). We modify the framework to support the training and evaluation pipelines proposed in this paper. For the training process, we utilize the \texttt{sglang}\footnote{\url{https://github.com/sgl-project/sglang}} framework as the rollout backend. Specifically, we modify the source code of \texttt{sglang} (version 0.4.6.post5) to implement the Semantic Neighbor Mixing logic described in the main text. Corresponding adaptations are made to the \texttt{verl} interface to ensure compatibility and seamless integration during the rollout phase.

\subsection{Hyperparameters for Training}

\begin{table}[h]
    \centering
    \begin{tabular}{lc}
        \toprule
        \textbf{Hyperparameter} & \textbf{Value} \\
        \midrule
        Number of Neighbors ($k$) & 3 \\
        Temperature ($\tau$) & 0.7 \\
        Global Batch Size & 64 \\
        Gradient Accumulation Step & 4 \\
        Max Prompt Length & 4096 \\
        Max Response Length & 8192 \\
        Learning Rate & 1e-6 \\
        KL Loss Coefficient ($\beta$) & 0.001 \\
        Group Size & 4 \\
        Epochs & 1 \\
        \bottomrule
    \end{tabular}
    \caption{\textbf{Hyperparameter settings used in the experiments.}}
    \label{tab:training_hyperparams}
\end{table}

We perform full-parameter fine-tuning on both DeepSeek-R1-Distill-Qwen-1.5B and DeepSeek-R1-Distill-Qwen-7B models for our main experiments. The detailed hyperparameters used during the training process are listed in Table~\ref{tab:training_hyperparams}.

\subsection{Hyperparameters for Inference}

For inference, we also employ the \texttt{sglang} engine. We adhere to the recommended settings for the base model, setting the sampling temperature to $0.6$ and nucleus sampling (top-$p$) to $0.95$. To maintain consistency with the training configuration, the maximum generation length is set to $8192$.

\section{Evaluation}
\label{sec:evaluation}

\subsection{Datasets}

We evaluate the reasoning capabilities of our models across four widely recognized benchmarks, categorized into in-domain mathematical reasoning and out-of-distribution scientific reasoning. The statistics of these datasets are summarized in Table~\ref{tab:datasets}.

\paragraph{Mathematical Reasoning.}
We utilize the AMC23, AIME24 and AIME25 to assess the model's proficiency in competition-level mathematics, with AIME serving as a more challenging holdout set requiring complex multi-step reasoning. Additionally, we use MATH500, a widely adopted subset of the MATH dataset~\cite{hendrycks2021measuring}, covering diverse subjects such as algebra, geometry, number theory, and probability.

\paragraph{Scientific Reasoning.}
To evaluate generalization beyond pure mathematics, we use GPQA-Diamond, a subset of the GPQA benchmark~\cite{rein2024gpqa} containing high-difficulty expert-level questions in biology, physics, and chemistry.

\begin{table}[h]
\centering
\small
\setlength{\tabcolsep}{12pt}
\begin{tabular}{@{}llc@{}}
\toprule
\textbf{Dataset} & \textbf{Domain} & \textbf{Size} \\ \midrule
AMC 23 & Math (Competition) & 40 \\
AIME 24 & Math (Competition) & 30 \\
AIME 25 & Math (Competition) & 30 \\
MATH500 & Math (General) & 500 \\
GPQA-Diamond & Science & 198 \\ \bottomrule
\end{tabular}
\caption{\textbf{Statistics of the evaluation benchmarks used in our experiments.}}
\label{tab:datasets}
\end{table}

\subsection{Metrics}

Following standard practices in reasoning evaluation, we report \textit{Mean@32} (average accuracy) and \textit{Pass@k} metrics. For each problem, we generate $n=32$ candidate solutions.

\paragraph{Mean@n.} 
This metric represents the expected accuracy of a single generation sampled from the model. For a set of $n$ generated samples, it is calculated as:
\begin{equation}
    \text{Mean@}n = \frac{1}{n} \sum_{i=1}^{n} \mathbb{I}(o_i \text{ is correct}),
\end{equation}
where $o_i$ denotes the $i$-th generated solution, and $\mathbb{I}(\cdot)$ is the indicator function. In our experiments, we compute this over $n=32$ samples. In Table~\ref{tab:math}, this metric is reported in the ``@1'' column, reflecting the expected performance of a single inference pass (an unbiased estimator of Pass@1).

\paragraph{Pass@k.}
We report Pass@16 and Pass@32 to evaluate the exploration potential of the model. Pass@$k$ estimates the probability that at least one correct solution exists within $k$ generated samples. Since we generate $n=32$ samples total, for $k \le n$, we calculate the unbiased estimator:
\begin{equation}
    \text{Pass@}k = 1 - \mathbb{E}\left[ \frac{\binom{n-c}{k}}{\binom{n}{k}} \right],
\end{equation}
where $c$ is the number of correct samples among the $n$ generations. If $n-c < k$, the probability is $1$. For Pass@32 (where $k=n$), this simplifies to checking if $c > 0$.

\subsection{Answer Extraction and Grading}

To ensure fair evaluation, we employ a strict rule-based extraction and grading pipeline.

\paragraph{Mathematical Tasks.}
For datasets requiring a numerical or symbolic answer (AMC, AIME, MATH500), the model is instructed to place the final answer within a \texttt{\textbackslash boxed\{\}} command. We extract the content inside the last boxed environment. The extracted answer is compared against the ground truth using the symbolic verification utilities provided by \texttt{Math-Verify}\footnote{\url{https://github.com/huggingface/Math-Verify}}. This allows for robust equivalence checks (e.g., matching fractions, decimals, and simplified expressions) to account for formatting variations.

\paragraph{Multiple Choice Tasks.}
For GPQA-Diamond, we parse the model output for the specific pattern "Answer: <Option>". The extracted option (A, B, C, or D) is directly compared to the reference label. If the model fails to follow the format or produces an ambiguous answer, it is marked as incorrect.

\section{Transfer to GSPO (N-GSPO)}
\label{app:ngspo}

To test whether the benefit of Semantic Neighbor Mixing is specific to the GRPO advantage-estimation recipe, we plug the same mixing module into GSPO, denoted \textbf{N-GSPO}, and evaluate on DeepSeek-R1-Distill-Qwen-1.5B. All other training and evaluation settings follow Section~\ref{ssec:implementation_details}. As reported in Table~\ref{tab:gspo}, N-GSPO improves the average Pass@32 over GSPO, with a particularly large $+7.66$ gain on AIME25. This demonstrates that semantically grounded embedding-level exploration during rollout transfers to other rollout-based optimization pipelines, and is not an artifact of GRPO-specific advantage estimation.

\begin{table}[h]
\small
\centering
\setlength{\tabcolsep}{4pt}
\begin{tabular}{@{}l|cccc@{}}
\toprule
 & AIME25 & AMC23 & MATH500 & Avg \\ \midrule
GSPO & 43.42 & \textbf{97.00} & \textbf{91.59} & 77.34 \\
N-GSPO & \textbf{51.08} & 94.87 & 91.17 & \textbf{79.04} \\ \bottomrule
\end{tabular}
\caption{\textbf{Transfer to GSPO.} Applying Semantic Neighbor Mixing to GSPO (N-GSPO) improves average Pass@32 on DeepSeek-R1-Distill-Qwen-1.5B.}
\label{tab:gspo}
\end{table}

\section{Semantic Coherence and Trajectory Diversity of Mixed Embeddings}
\label{app:coherence_diversity}

A central design question for \Method is whether the mixed embedding is \emph{close enough} to the anchor to preserve semantic coherence, yet \emph{different enough} to actually steer the rollout into new reasoning branches. We examine both ends of this trade-off.

\paragraph{Proximity to the anchor semantics.}
We measure the cosine similarity between the mixed embedding $\tilde{e}_{t+1}$ and the anchor token embedding $E[v_t^*]$ over all mixed steps during rollout on DeepSeek-R1-Distill-Qwen-1.5B. We obtain an average cosine similarity of $0.9985$ with a minimum of $0.8813$, and only $1.2\%$ of mixed steps fall below $0.95$. This confirms that Semantic Neighbor Mixing, by construction, keeps the input representation tightly within the local semantic neighborhood of the anchor. 

\paragraph{Meaningful trajectory variation.}
To demonstrate that such tightly-constrained mixing still induces non-trivial trajectory changes, we conduct a paired inference-time experiment on AIME24 with DeepSeek-R1-Distill-Qwen-1.5B. Keeping all hyperparameters identical to our main study, we generate $32$ samples per problem under two decoding strategies, \textit{Standard} temperature sampling and \textit{Mixing} (enabling Semantic Neighbor Mixing at inference), and compare their per-problem Pass@32 outcomes in Table~\ref{tab:mixing_paired}.

\begin{table}[h]
\small
\centering
\setlength{\tabcolsep}{6pt}
\begin{tabular}{@{}l|cc@{}}
\toprule
 & Mixing Correct & Mixing Wrong \\ \midrule
Standard Correct & 63.3\% & 0.0\% \\
Standard Wrong & \textbf{10.0\%} & 26.7\% \\ \bottomrule
\end{tabular}
\caption{\textbf{Paired Pass@32 outcomes on AIME24} between standard sampling and Semantic Neighbor Mixing. Mixing uniquely solves $10.0\%$ of the problems that standard sampling fails, with no problem solved by standard-only.}
\label{tab:mixing_paired}
\end{table}

Strikingly, Mixing uniquely resolves $10.0\%$ of the problems that are unsolved by standard sampling, and does not lose any problem that was solvable under standard sampling. This confirms that the modest but non-zero perturbation introduced by Semantic Neighbor Mixing is sufficient to steer the autoregressive process into genuinely new reasoning branches, which is exactly the behavior desired during RL rollout exploration.

\section{Full Ablation Results}
\label{app:full_ablation_results}

In this section, we provide the comprehensive numerical results for the ablation studies discussed in the main text. These tables offer a detailed breakdown of performance across all evaluated metrics for the individual benchmarks (AIME25, AMC23, and MATH500). Note that in the table headers, the column labeled \textbf{@1} denotes the \textbf{Mean@32} (average accuracy), while @16 and @32 correspond to Pass@16 and Pass@32, respectively.

Table~\ref{tab:exploration_rate} presents the sensitivity analysis for the mixing rate parameter $\rho$ across both the 1.5B and 7B model scales.

Table~\ref{tab:mix_strategy_full} details the impact of different mixing strategies on the 1.5B model.

Table~\ref{tab:distance_measurement_full} compares the efficacy of different distance metrics for neighbor selection using the DeepSeek-R1-Distill-Qwen-1.5B model.

Lastly, Table~\ref{tab:inference_strategy_full} presents a granular analysis of the inference strategies discussed in Section 4.5. We compare the performance of models trained with \Method when evaluated using standard temperature sampling versus identifying and mixing semantic neighbors at test time.

\begin{table*}[ht]
\small
\centering
\setlength{\tabcolsep}{6pt}
\begin{tabular}{@{}lcccccccccccc@{}}
\toprule
\multicolumn{1}{c|}{\multirow{2}{*}{Mixing Rate}} & \multicolumn{3}{c|}{\textbf{AIME25}} & \multicolumn{3}{c|}{\textbf{AMC23}} & \multicolumn{3}{c|}{\textbf{MATH500}} & \multicolumn{3}{c}{\textbf{Average}} \\
\multicolumn{1}{c|}{} & @1 & @16 & \multicolumn{1}{c|}{@32} & @1 & @16 & \multicolumn{1}{c|}{@32} & @1 & @16 & \multicolumn{1}{c|}{@32} & @1 & @16 & @32 \\ \midrule
\multicolumn{13}{c}{\textit{\textbf{DeepSeek-R1-Distill-Qwen-1.5B}}} \\ \midrule
\multicolumn{1}{l|}{0.1} & 23.33 & \textbf{46.48} & \multicolumn{1}{c|}{\textbf{50.28}} & 64.14 & 94.07 & \multicolumn{1}{c|}{96.18} & 74.87 & \textbf{89.92} & \multicolumn{1}{c|}{\textbf{91.05}} & 54.11 & \textbf{76.82} & \textbf{79.17} \\
\multicolumn{1}{l|}{0.15} & \textbf{23.96} & 45.12 & \multicolumn{1}{c|}{48.15} & 64.38 & 93.51 & \multicolumn{1}{c|}{95.60} & 75.23 & 89.67 & \multicolumn{1}{c|}{90.62} & \textbf{54.52} & 76.10 & 78.12 \\
\multicolumn{1}{l|}{0.2} & 21.35 & 41.37 & \multicolumn{1}{c|}{45.69} & \textbf{64.84} & \textbf{94.93} & \multicolumn{1}{c|}{\textbf{96.49}} & \textbf{75.24} & 89.54 & \multicolumn{1}{c|}{90.78} & 53.81 & 75.28 & 77.65 \\ \midrule
\multicolumn{13}{c}{\textit{\textbf{DeepSeek-R1-Distill-Qwen-7B}}} \\ \midrule
\multicolumn{1}{l|}{0.1} & \textbf{34.27} & 58.00 & \multicolumn{1}{c|}{61.18} & 80.70 & \textbf{96.92} & \multicolumn{1}{c|}{\textbf{98.13}} & \textbf{84.94} & \textbf{92.77} & \multicolumn{1}{c|}{\textbf{93.28}} & \textbf{66.64} & \textbf{82.56} & \textbf{84.20} \\
\multicolumn{1}{l|}{0.15} & 33.44 & 56.56 & \multicolumn{1}{c|}{59.02} & \textbf{81.72} & 94.55 & \multicolumn{1}{c|}{95.76} & 84.74 & 92.34 & \multicolumn{1}{c|}{92.90} & 66.63 & 81.15 & 82.56 \\
\multicolumn{1}{l|}{0.2} & 33.96 & \textbf{58.81} & \multicolumn{1}{c|}{\textbf{61.58}} & 79.77 & 95.89 & \multicolumn{1}{c|}{97.83} & 84.57 & 92.48 & \multicolumn{1}{c|}{93.17} & 66.10 & 82.39 & 84.19 \\ \bottomrule
\end{tabular}
\caption{\textbf{Detailed ablation study on the mixing rate parameter ($\rho$) for DeepSeek-R1-Distill-Qwen-1.5B and 7B models.} We evaluate performance using Mean@32 (@1), Pass@16 (@16), and Pass@32 (@32) across AIME25, AMC23, and MATH500.}
\label{tab:exploration_rate}
\end{table*}

\begin{table*}[ht]
\small
\centering
\setlength{\tabcolsep}{4pt}
\begin{tabular}{@{}l|ccc|ccc|ccc|ccc@{}}
\toprule
 & \multicolumn{3}{c|}{\textbf{AIME25}} & \multicolumn{3}{c|}{\textbf{AMC23}} & \multicolumn{3}{c|}{\textbf{MATH500}} & \multicolumn{3}{c}{\textbf{Average}} \\
 & @1 & @16 & @32 & @1 & @16 & @32 & @1 & @16 & @32 & @1 & @16 & @32 \\ \midrule
\textit{DeepSeek-R1-Distill-Qwen-1.5B} & 20.83 & 38.05 & 41.19 & \textbf{64.14} & 91.73 & 92.37 & 74.88 & 89.38 & 90.29 & 53.28 & 73.05 & 74.62 \\
+Gumbel Soft-Thinking & 22.40 & 42.33 & 46.49 & 63.98 & 91.88 & 93.61 & \textbf{75.24} & 89.52 & 90.58 & 53.87 & 74.58 & 76.89 \\
+\Method w/o rate & 22.71 & 42.10 & 45.21 & 62.81 & 93.62 & 95.91 & 75.16 & 89.76 & 90.83 & 53.56 & 75.16 & 77.32 \\
+\Method & \textbf{23.33} & \textbf{46.48} & \textbf{50.28} & \textbf{64.14} & \textbf{94.07} & \textbf{96.18} & 74.87 & \textbf{89.92} & \textbf{91.05} & \textbf{54.11} & \textbf{76.82} & \textbf{79.17} \\ \bottomrule
\end{tabular}
\caption{\textbf{Comprehensive performance comparison of mixing strategies on the DeepSeek-R1-Distill-Qwen-1.5B model.} We compare \Method against the Gumbel Soft-Thinking baseline and an unconstrained variant (w/o rate) where the mixing mechanism is applied to all tokens.}
\label{tab:mix_strategy_full}
\end{table*}

\begin{table*}[ht]
\small
\centering
\setlength{\tabcolsep}{5pt}
\begin{tabular}{@{}l|lll|lll|lll|lll@{}}
\toprule
\multicolumn{1}{c|}{\multirow{2}{*}{Distance Measurement}} & \multicolumn{3}{c|}{\textbf{AIME25}} & \multicolumn{3}{c|}{\textbf{AMC23}} & \multicolumn{3}{c|}{\textbf{MATH500}} & \multicolumn{3}{c}{\textbf{Average}} \\
\multicolumn{1}{c|}{} & \multicolumn{1}{c}{@1} & \multicolumn{1}{c}{@16} & \multicolumn{1}{c|}{@32} & \multicolumn{1}{c}{@1} & \multicolumn{1}{c}{@16} & \multicolumn{1}{c|}{@32} & \multicolumn{1}{c}{@1} & \multicolumn{1}{c}{@16} & \multicolumn{1}{c|}{@32} & \multicolumn{1}{c}{@1} & \multicolumn{1}{c}{@16} & \multicolumn{1}{c}{@32} \\ \midrule
Cosine & \textbf{23.33} & \textbf{46.48} & \textbf{50.28} & \textbf{64.14} & \textbf{94.07} & \textbf{96.18} & 74.87 & \textbf{89.92} & \textbf{91.05} & \textbf{54.11} & \textbf{76.82} & \textbf{79.17} \\
L2 & 22.29 & 43.49 & 47.13 & 63.83 & 92.71 & 95.24 & 75.18 & 89.57 & 90.67 & 53.77 & 75.26 & 77.68 \\
L1 & 20.83 & 38.48 & 41.93 & \textbf{63.36} & 90.78 & 92.83 & \textbf{75.02} & 89.66 & 90.73 & 53.07 & 72.98 & 75.16 \\ \bottomrule
\end{tabular}
\caption{\textbf{Impact of different distance metrics on latent space neighbor selection for the DeepSeek-R1-Distill-Qwen-1.5B model.} We report results using Cosine distance, L2 (Euclidean) distance, and L1 (Manhattan) distance.}
\label{tab:distance_measurement_full}
\end{table*}

\begin{table*}[ht]
\small
\centering
\setlength{\tabcolsep}{4pt}
\begin{tabular}{@{}lcccccccccccc@{}}
\toprule
\multicolumn{1}{c|}{\multirow{2}{*}{Inference Strategy}} & \multicolumn{3}{c|}{\textbf{AIME25}} & \multicolumn{3}{c|}{\textbf{AMC23}} & \multicolumn{3}{c|}{\textbf{MATH500}} & \multicolumn{3}{c}{\textbf{Average}} \\
\multicolumn{1}{c|}{} & @1 & @16 & \multicolumn{1}{c|}{@32} & @1 & @16 & \multicolumn{1}{c|}{@32} & @1 & @16 & \multicolumn{1}{c|}{@32} & @1 & @16 & @32 \\ \midrule
\multicolumn{13}{c}{\textit{\textbf{\Method trained DeepSeek-R1-Distill-Qwen-1.5B}}} \\ \midrule
\multicolumn{1}{l|}{Standard} & \textbf{23.33} & \textbf{46.48} & \multicolumn{1}{c|}{\textbf{50.28}} & \textbf{64.14} & \textbf{94.07} & \multicolumn{1}{c|}{\textbf{96.18}} & 74.87 & \textbf{89.92} & \multicolumn{1}{c|}{\textbf{91.05}} & \textbf{54.11} & \textbf{76.82} & \textbf{79.17} \\
\multicolumn{1}{l|}{Semantic Neighbor Mixing} & 21.98 & 41.13 & \multicolumn{1}{c|}{44.88} & 62.58 & 93.13 & \multicolumn{1}{c|}{95.52} & \textbf{75.01} & 89.76 & \multicolumn{1}{c|}{90.76} & 53.19 & 74.67 & 77.05 \\ \midrule
\multicolumn{13}{c}{\textit{\textbf{\Method trained DeepSeek-R1-Distill-Qwen-7B}}} \\ \midrule
\multicolumn{1}{l|}{Standard} & \textbf{34.27} & \textbf{58.00} & \multicolumn{1}{c|}{\textbf{61.18}} & \textbf{80.70} & \textbf{96.92} & \multicolumn{1}{c|}{\textbf{98.13}} & 84.94 & \textbf{92.77} & \multicolumn{1}{c|}{\textbf{93.28}} & \textbf{66.64} & \textbf{82.56} & \textbf{84.20} \\
\multicolumn{1}{l|}{Semantic Neighbor Mixing} & 31.88 & 51.56 & \multicolumn{1}{c|}{54.52} & 80.16 & 95.69 & \multicolumn{1}{c|}{96.61} & \textbf{85.04} & 92.75 & \multicolumn{1}{c|}{93.22} & 65.69 & 80.00 & 81.45 \\ \bottomrule
\end{tabular}
\caption{\textbf{Detailed performance comparison of inference strategies.} We evaluate the impact of enabling Semantic Neighbor Mixing during the inference phase versus using standard temperature sampling on models trained with \Method. The results cover both 1.5B and 7B model scales across all three mathematical reasoning benchmarks.}
\label{tab:inference_strategy_full}
\end{table*}

\section{Rollout Throughput Overhead}
\label{app:throughput}

To quantify the practical cost of introducing Semantic Neighbor Mixing into the rollout loop, we measure the average rollout throughput (in tokens/s) on identical hardware with the same rollout backend. \textit{Standard Sampling} corresponds to GRPO rollouts (no mixing) and \textit{Semantic Neighbor Mixing} corresponds to \Method rollouts (mixing enabled). We report the relative slowdown:
\[
\mathrm{Slowdown}(\%) =
\left(1 - \frac{\mathrm{Throughput}_{\mathrm{mix}}}{\mathrm{Throughput}_{\mathrm{std}}}\right) \times 100.
\]

As shown in Table~\ref{tab:throughput}, the overhead is modest (below $10\%$) at both 1.5B and 7B scales. This is expected because (i) Semantic Neighbor Mixing activates on only a small fraction of tokens (controlled by the mixing rate $\rho$), (ii) the top-$k$ nearest-neighbor sets can be precomputed once the policy is frozen during the rollout of one iteration, and (iii) the weighted aggregation is a small weighted sum over $k$ neighbor embeddings. In exchange for this minor overhead, \Method delivers consistent accuracy gains on math reasoning, improving the quality--efficiency trade-off.

\begin{table*}[t]
\small
\centering
\begin{tabular}{@{}lccc@{}}
\toprule
 & Std.\ (tok/s) & Mixing (tok/s) & slowdown\% \\ \midrule
DS-R1-Distill-Qwen-1.5B & 1362.88 & 1232.00 & 9.60 \\
DS-R1-Distill-Qwen-7B & 663.62 & 604.54 & 8.90 \\ \bottomrule
\end{tabular}
\caption{\textbf{Rollout throughput comparison} between standard sampling (GRPO) and Semantic Neighbor Mixing (\Method).}
\label{tab:throughput}
\end{table*}

\end{document}